\documentclass{article}
\usepackage{multirow}
\usepackage[normalem]{ulem}
\useunder{\uline}{\ul}{}
\usepackage{amsmath}
\usepackage{PRIMEarxiv}
\usepackage[utf8]{inputenc} 
\usepackage[T1]{fontenc}    
\usepackage{hyperref}       
\usepackage{url}            
\usepackage{booktabs}       
\usepackage{amsfonts}       
\usepackage{nicefrac}       
\usepackage{microtype}      
\usepackage{lipsum}
\usepackage{fancyhdr}       
\usepackage{graphicx}       
\graphicspath{{media/}}     

\title{SDNet: Multi-branch for single Image Deraining using Swin}

\author{
  Fuxiang Tan \\
  School of Software,Xinjiang University \\
  \texttt{tanfuxiang5018@stu.xju.edu.cn} \\
   \And
  YuTing Kong \\
  School of Software,Xinjiang University \\
  \texttt{kongyuting1175@stu.xju.edu.cn} \\
  \And
  Yingying Fan \\
  School of Software,Xinjiang University \\
  \And
  Feng Liu \\
  School of Software,Xinjiang University \\
  \And
  Daxin Zhou \\
  School of Software,Xinjiang University \\
  \And
  Hao zhang \\
  School of Software,Xinjiang University \\
  \And
  Long Chen \\
  School of Software,Xinjiang University \\
  \And
  Liang Gao \\
  School of Software,Xinjiang University \\
  \And
  Yurong Qian\thanks{Corresponding author} \\
  School of Software,Xinjiang University \\
  \texttt{qyr@xju.edu.cn} \\

}

\begin{document}
\maketitle

\begin{abstract}
    Rain streaks degrade the image quality and seriously affect the performance of subsequent computer vision tasks, such as autonomous driving, social security, etc. Therefore, removing rain streaks from a given rainy images is of  great significance. Convolutional neural networks(CNN) have been widely used in image deraining tasks, however, the local computational characteristics of convolutional operations limit the development of image deraining tasks. Recently, the popular transformer has global computational features that can further facilitate  the development of image deraining tasks. In this paper, we introduce Swin-transformer into the field of image deraining for the first time to study the performance and potential of Swin-transformer in the field of image deraining. Specifically, we improve the basic module of Swin-transformer and design a three-branch model to implement single-image rain removal. The former implements the basic rain pattern feature extraction, while the latter fuses different features to further extract and process the image features. In addition, we employ a jump connection to fuse deep features and shallow features. In terms of experiments, the existing public dataset suffers from image duplication and relatively homogeneous background. So we propose a new dataset Rain3000 to validate our model. Therefore, we propose a new dataset Rain3000 for validating our model. Experimental results on the publicly available datasets Rain100L, Rain100H and our dataset Rain3000 show that our proposed method has performance and inference speed advantages over the current mainstream single-image rain streaks removal models.The source code will be available at  
    \url{https://github.com/H-tfx/SDNet}.
\end{abstract}


\section{Introduction}
\label{introduction}
   At present, target detection and image segmentation are mostly performed on images or videos with good shooting environment. Bad weather can degrade the quality of captured content and limit the use scenarios and performance of other computer vision tasks such as target detection, and rain is typical of bad weather. Compared with video, the lack of temporal information in images can lead to more complex and difficult single-image deraining. In addition, most deraining tasks are still image-based due to the sparsity of information density of video contents and the high storage cost, so the research of single-image deraining has received wide attention. Currently, most studies consider that the rain pattern image's are formed by adding rain streaks to the background image with the following equation:
   \begin{equation}\label{eq1}   \mathrm{Y}=\mathrm{B}+\mathrm{R}   \end{equation}
   where Y is the rainy image, B is the deraining image, and R is the rain streaks image. According to Equation \ref{eq1}, the previous rain removal methods can be roughly divided into two categories, model-driven methods\cite{ref1,ref2,ref3,ref4,ref5,ref6} and data-driven methods\cite{ref7,ref8,ref9,ref10,ref11,ref12,ref13,ref14,ref15}.Before 2017, it was mostly model-driven. After 2017, data-driven (a deep learning method) has gradually become the mainstream of image removal\cite{ref16}. Model-driven methods are mainly based on image decomposition\cite{ref3,ref17}, sparse representation\cite{ref1} low-rank representation\cite{ref27} and Gaussian mixture\cite{ref4} and other models. These models are mostly based on a priori knowledge of the design of artificial methods to extract features, although these methods have made some achievements, but there are still large room for improvement.
    
    In recent years, the data-driven CNN method has been significantly better than the model-driven method. Most CNN-based deep learning methods constrain the network according to Equation 1. For example, Liu et al. \cite{ref17} construct a decomposition network to decompose the input image into a deraining image and a rain streaks image. The obtained rain streaks image can be used to construct cyclic consistency. The loss function further constrains the network. In order to obtain rain streaks image from a multi-scale perspective, Wangle et al. \cite{ref18} used multi-stage dilated convolution to obtain different receptive fields, and used the layers with smaller receptive fields to guide the learning of larger receptive fields. Better residual learning. Single-image rain removal is a kind of image restoration research. Deng et al.\cite{ref19} proposed to divide rain removal and detail restoration into two independent tasks, so that the function of each part can be fully utilized. The first part is the mapping of the input image to the rain streaks image, and then the rain streaks image is subtracted from the input image to obtain a rougher deraining image. The second part is used to extract high-frequency detail information to repair the rough background image. The above are all based on supervision. In addition, there are semi-supervised \cite{ref20,ref21,ref22,ref34} 
    and unsupervised \cite{ref35,ref36,ref37}.
    These deep learning methods all have a problem of low feature propagation efficiency over long distances. Most of them use jump connections to improve the efficiency of feature propagation, but this will cause the problem of relatively difficult extraction of deep features.Another limitation is that the convolution operation of shared weights can only focus on local features. If you want to focus on global information, you can only stack convolution blocks.
    
    Transformer \cite{ref23} was originally a model in the field of natural language processing (NLP) for parallel processing of word vectors to accelerate model inference. Its global computational properties are suitable for passing features over long distances. This is exactly what the convolutional operations in the field of computer vision are not good at. Dosovitskiy et al\cite{ref24} sliced the image into 16x16 image blocks, and different image blocks were considered as different words and input to the transformer, which improved the accuracy of image classification. Recently there are applications of transformer from depth \cite{ref25}, multi-scale \cite{ref26} and other perspectives to accomplish related tasks. However, Transformer also has non-negligible disadvantages, such as a quadratic relationship between computation and image size, which can limit its application environment. The Swin-transformer proposed by Liu et al \cite{ref29} uses sliding windows to make the model with linear computational complexity, improves the information exchange between windows with cross-window connections, and finally improves the performance of the model in image classification, target detection, and instance segmentation. 
    
    In this paper, we propose a new image de-rainage network, called SDNet, which is an end-to-end de-rainage network built using the powerful feature representation capability of Swin-transformer. Specifically, we improve the basic module of Swin-transformer and redesign a two-branch model to implement single-image deraining. The former implements the basic rain pattern feature extraction, while the latter fuses the features of different branches. In addition, we employ jump connections to fuse deep features and shallow features to improve the performance of the network model. Numerous experiments demonstrate the effectiveness of our proposed network in the image deraining task.
    
    In summary, the contributions we have made are as follows:
        \begin{itemize}
        \item We have improved the Swin-transformer module to make the output size of the basic block consistent with the input for feature extraction.
        \item We propose a three-branch network module to improve the efficiency of feature fusion and use additional jump connections to fuse deep and shallow features to improve the learning ability of the network.
        \item We propose a completely new dataset. And we use this dataset to train and test our proposed model as well as other mainstream rain removal models.
        \item Experiments demonstrate that our model has some performance and inference speed advantages over the mainstream deraining model.
        \end{itemize}

\section{Related work}
\label{related work}

    In this section, we briefly review the development of the single-image deraining algorithm model and some applications of transformer in the CV field, and then lead to our approach.

    \subsection{Deraining}
    The image deraining task can usually be expressed as obtaining a mapping from an image with rain to an image without rain. The mappings can be broadly classified into model-driven and data-driven approaches based on the method used to obtain them. The model-based drive is designed based on a priori knowledge of rain pattern images. The literature \cite{ref27} designed a low-rank representation based on the fact that rain streaks in the same image have a similar or identical shape, giving the image a low rank. The literature \cite{ref6} proposes a method for image deraining through a sparse discriminant dictionary, starting from the sparsity of rain pattern features. In some rain images, the rain pattern is more similar to the background, which can lead to insufficient or excessive derainting, and can even cause the background texture to smooth out. To solve this problem, a Gaussian mixture model is designed for image deraining by proposing a method to complement the image with similar blocks in the literature \cite{ref4} . The literature \cite{ref2} decomposes the image into high and low frequencies by means of image decomposition, and then uses sparse coding for high-frequency information to remove rain patterns and retain background detail information. The literature \cite{ref1} constructs a global sparse model containing three sparse terms for the image deraining task, starting from the knowledge background of the inherent morphology of rain patterns.
    
    Although the model-driven methods have achieved good results, they all have the problem of weak generalization ability and the effect of rain removal still need to be improved. Recently, with the increase of hardware computing power, the data-driven deep learning side has achieved impressive results. The method is to learn a nonlinear mapping function from a rain map to a background image using paired data sets. DerainNet proposed by Fu et al \cite{ref28} is the first known model to apply a deep learning approach to the image rain removal task, which only makes high frequency information to train the network.Du et al \cite{ref30} considered that rain patterns are different at different spatial locations and color channels and proposed a conditional variational rain removal network, while a rain pattern density estimation module was proposed for the purpose of adaptive rain pattern density. for estimating the rain pattern density.Zhang Zhang et al. \cite{ref31} also designed a multi-stream density estimator from the perspective of density to realize the adaptive rain removal of the image. He et al. \cite{ref32} jointly considered rain pattern density and raindrop size, and proposed a multi-scale rain pattern density estimation module to guide the network to remove rain. Jiang et al. \cite{ref33} further studied the effect of multi-scale models on rain removal and proposed a multi-scale progressive fusion model. Wang et al. \cite{ref7} also noticed the importance of multi-scale information for rain removal tasks, and proposed a scale aggregation module to learn features of different scales. In addition, a self-attention module was introduced to make feature aggregation adapt to each channel. Whether it is strong supervision, semi-supervision  \cite{ref20,ref21,ref22,ref34}, or unsupervised \cite{ref35,ref36,ref37}
    , a fixed convolution kernel sharing parameter method is used. This method reduces the amount of parameters, but also limits the receptive field of the convolution kernel and the ability to adapt to different scales. Although technologies such as multi-scale and multi-core stacking have been used to alleviate these problems, these problems have not been fundamentally solved. In addition, the new technology introduced has brought other problems. For example, the number of stacking is too much, which hinders the transfer of features.
    
    \subsection{Transformer}
    There has been a large amount of recent research work to introduce Transformer into the CV domain with good results. Specifically, Dosovitskiy et al \cite{ref24} divided images into 16X16 image blocks and then stretched them into one-dimensional vectors that were fed into a network to accomplish the task of image classification. Chen et al \cite{ref38} proposed TransUnet by combining transformer and Unet based on convolutional operations to achieve segmentation of medical images. Jiang et al \cite{ref39} designed the transformer with the same structure as the adversarial generative network for image generation. self-attention in the transformer causes the model computation to grow squarely, resulting in the transformer not being able to run on low computing power hardware. Liu \cite{ref29} proposed the Swin- transformer uses a sliding window approach to make the network computation grow linearly and speed up the inference of the network. Our approach is based on this method to implement a single image deraining task in terms of fusing features from different branches, deep features and shallow features. To the best of our knowledge, we are the first to build an image deraining task based on Swin-transformer.
    
    \section{Method}
    In this section, we present in detail the single image deraining network developed based on the Swin-transformer network. Specifically, first in Section 3.1, we briefly introduce the Swin-transformer, then in Section 3.2 we describe the general framework of our proposed network, the multi-branch deraining module, called MSwt, and finally in Section 3.3 we introduce the loss function we use.
    \subsection{Swin-transformer}
    Transformer is a powerful network module that may replace CNN operations. But the Multi-Head Attention in it leads to a rapid increase in the computation of the model, which leads to the transformer model cannot be tested and used in many low level hardware.The mathematical expression of Attention is as follows:
        
        \begin{equation}
        \label{eq2}
        \operatorname{Attention}(Q, K, V)=\operatorname{SoftMax}\left(Q K^{T} / \sqrt{d}+B\right) V
        \end{equation}
        
        \begin{figure}
        \centering
        \includegraphics[scale=0.6]{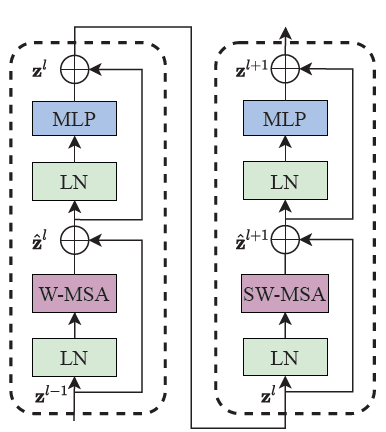}
        \caption{Sample figure caption.}
        \label{fig1}
        \end{figure}
    
     Where $Q, K, V \in \mathbb{R}^{M^{2} \times d}$ are the query, key and value matrices.$M^{2}$ represents the number of patches in a single Windows. d represents the dimension of query or key. The values in B are taken from the bias matrix $\hat{B} \in \mathbb{R}^{(2 M-1) \times(2 M-1)}$.  Swin-transformer proposes to limit the amount of parameters involved in a single self-attention calculation by fixing the window size, and then realize the information exchange between windows by means of sliding windows. The Swin-transforme block flow is shown in Figure \ref{fig1}. Swin-transformer uses feature compression to obtain high-level semantic features and achieves the best results in image classification, target detection and image segmentation. Loss and cannot be recovered, so this model cannot be directly used to remove rain from a single image. As far as we know, we are the first team to explore the use of this module for a single image rain removal task.
    
    \subsection{Network structure}
    
        \begin{figure}
        \centering
        \includegraphics[scale=0.3]{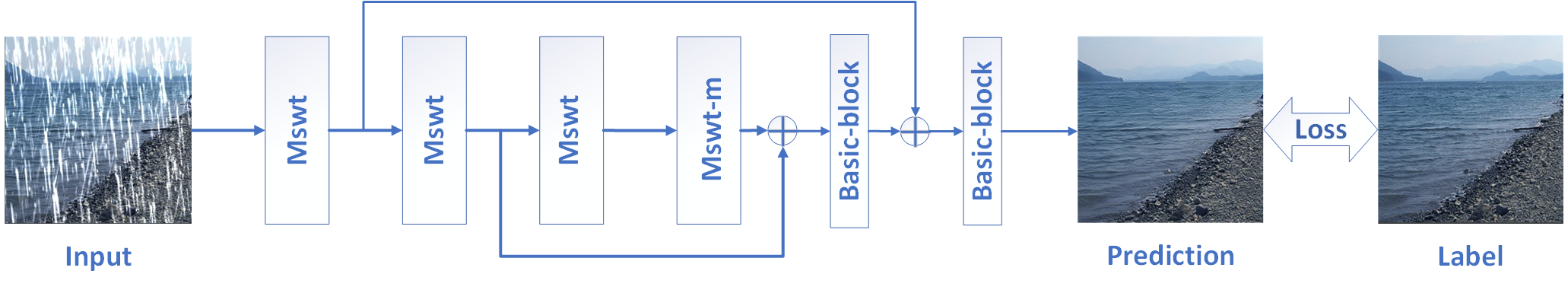}
        \caption{Sample figure caption.}
        \label{fig2}
        \end{figure}
    
    In this work, we utilize a simple and powerful feedforward network as the backbone, as shown in Figure \ref{fig2}. The SDnet network basically consists of three multi-branch fusion modules, called MSwt, one multi-branch module Mswt-m, and two Basic-bloack modules. In addition, we have added jump connections with the purpose of fusing deep and shallow features to improve the performance of network deraining.
    
        \begin{figure}
        \centering
        \includegraphics[scale=0.6]{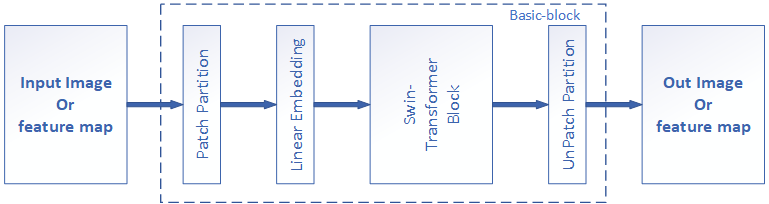}
        \caption{Sample figure caption.}
        \label{fig3}
        \end{figure}
        
    Inspired by Swin-transformer , our Basic-block module is designed as shown in Figure \ref{fig3}. The size of the input image or feature map is 231 × 231 × 3, and the size of the patch block is set to 3 × 3 in PatchPartition, so that the matrix size of the position encoding is $\frac{\mathrm{H}}{3} \times \frac{\mathrm{W}}{3}$.This is because more attention is paid to pixel information and location information in image de-rain tasks, while more attention is paid to semantic information in other advanced tasks such as image classification, so the patch blocks of both VIT \cite{ref24} and literature \cite{ref40} are set to 16 × 16 and the size of the location encoding is $\frac{\mathrm{H}}{16} \times \frac{\mathrm{W}}{16}$. The elongated vector is mapped to two times its length with full concatenation in Linear Embedding to avoid feature compression. In Swin-transformer block the window size is set to 7 and the number of heads of self-attention is set to 3. We add an Unpatch Partition operation after the Swin-transformer block, which is a reciprocal operation with Patch Partition to make the size and dimension of the output image or feature the same as the input one. In Swin-transformer Block W-MSA and SW-MSA together complete a global feature extraction. In our work, Basic-block uses a set.
    
\begin{figure}[htbp]
        \begin{minipage}[t]{0.5\linewidth}
        \centering
        \includegraphics[height=5.5cm,width=7cm]{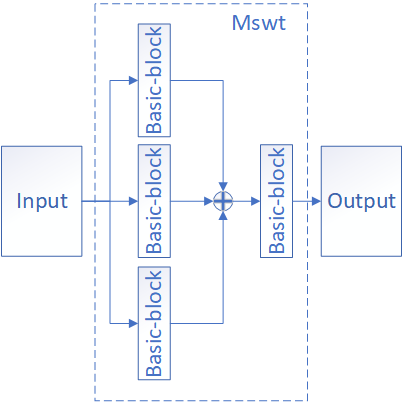}
        \caption{Mswt}
        \label{fig4}
        \end{minipage}%
        \hfill
        \begin{minipage}[t]{0.5\linewidth}
        \centering
        \includegraphics[height=5.5cm,width=7cm]{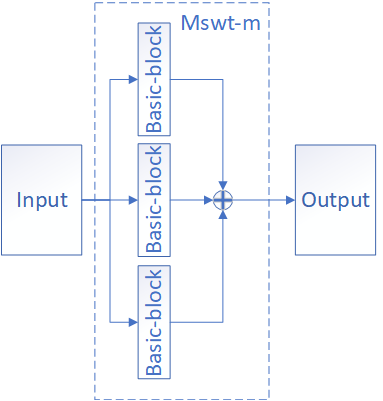}
        \caption{Mswt-m}
        \label{fig5}
        \end{minipage}
        \end{figure}

    The Basic-Block block is proposed to build the network more flexibly, and in this work, we designed two three-branch feature fusion blocks. As shown in Figure \ref{fig4} and Figure \ref{fig5}, the former has an additional Basic-Block block for fusing features compared to the latter. The mathematical expressions are as follows:
        \begin{equation}
        \label{eq3}
        f_{\mathrm{Msmt}}=F_{4}\left(F_{1}(x)+F_{2}(x)+F_{3}(x)\right)
        \end{equation}
    Where $\mathrm{F}(\cdot)$ denotes the operation of the Basic-block module. x denotes the input of the module Mswt and is the output. The idea of this design is derived from the multi-head attention mechanism in self-attention. Through learning $\mathrm{F}_{1}$, $\mathrm{F}_{2}$, $\mathrm{F}_{3}$ it can adaptively learn different features. The input is mapped to different subspaces and different features are extracted separately. The difference with self-attention is that we sum the extracted features instead of concatenation operation. The added features are fused through $\mathrm{F}_{4}$ to achieve further feature extraction. Since the design idea comes from the multi-head attention mechanism, multi-branch has the same characteristics as this mechanism, that is, within a certain range, the more branches, the better the model performance. In order to balance the scale of the model and the model performance, we choose then three branches to extract the features.
    
    Although the transformer can keep the features propagated over long distances, there is still a need to combine deep and shallow features in the network, for which we designed a Mswt module without feature fusion, which we call Mswt-m, as shown in Figure 5, with the following mathematical expressions:
        \begin{equation}
        \label{eq4}
        f_{\mathrm{Mswt}-\mathrm{m}}=F_{1}(x)+F_{2}(x)+F_{3}(x)
        \end{equation}
    $\mathrm{F}_{1}$, $\mathrm{F}_{2}$, $\mathrm{F}_{3}$  map the input to three different spaces to extract features, sum the extracted features and then sum them with the output of the second Mswt module, and then go through a Basic-block to achieve the fusion of depth features and shallow features, as shown in the small jump connection in Figure \ref{fig2}. The relatively long jump connection in Figure \ref{fig2}, on the other hand, takes into account the rich spatial and texture information contained in the primary features, which helps to complete the missing texture information in the depth features. The effectiveness of the jump connections will be demonstrated in the ablation experiments in Section 4.2
    
    \subsection{Loss function}
    In this work, we use the expression of the loss function as shown in Equation 5:
        \begin{equation}
        \label{eq5}
        \text { Loss }_{\text {all }}=\alpha \times \text { Loss }_{L_{1}}+\beta \times\left(\text { 1-Loss }_{\text {SSIM }}\right)+\lambda \times \text { Loss }_{\text {Ide}}
        \end{equation}
        \begin{equation}
        \label{eq6}
        \text { Loss }_{\mathrm{L}}=\mathrm{L} 1(\mathrm{SDnet}(\mathrm{O}), \mathrm{B})
        \end{equation}
        \begin{equation}
        \label{eq7}
        \text { Loss }_{\text {ssIM }}=\text { SSIM }(\text { SDnet }(O), \text { B })
        \end{equation}
        \begin{equation}
        \label{eq8}
        \text { Loss }_{\text {ide }}=\mathrm{L} 1(\mathrm{SDnet}(\mathrm{B}), \mathrm{B})
        \end{equation}
    Where, O is the image with rain pattern and B is the corresponding label. is the Sum of AbsoluteDifference (SAD), which is used to calculate the pixel loss between like predicted images and labels, as shown in Equation \ref{eq6}. SSIM (Structural Similarity) is the structural similarity, originally used as a metric to evaluate the structural similarity of the content of two images. the negative of SSIM as The effectiveness of the loss function for the image deraining task has been demonstrated by Ren et al \cite{ref41}, and the mathematical expression is shown in Equation \ref{eq7}. Although a high SSIM metric can be obtained using this loss function, the image still suffers from distortion and low peak signal-to-noise ratio (PSNR). The identity loss (Equation \ref{eq8}) is derived from CycleGAN \cite{ref42} used to constrain the color loss of the generated images, here we use to constrain the image style after image de-rainage, which reduces the image distortion and improves the network performance. $\alpha$ , $\beta$ , $\lambda$   are the coefficients of SAD loss, SSIM loss and identity loss. In our research, set to 0.2, 4 and 1 respectively.
\section{Experimental Results}
\label{experimental}
    In this part, we conducted a large number of experiments on our proposed dataset and two public datasets to prove the effectiveness of our proposed method. Specifically, in the ablation experiment in Section 4.2, we will prove the effectiveness of the proposed model and determine specific hyperparameters. In Section 4.3, we compare the results with six advanced rain removal models to prove the advantages of the proposed method.
    
    \subsection{Experimental environment}
    We trained with a Tesla V100 16G GPU, using the framework Pytorch 1.7.0 and the classical adaptive moment estimation optimizer (Adam) \cite{ref43}.The initial value of the learning rate is $5 \times 10^{-4}$ , which is reduced to $5 \times 10^{-5}$ and $5 \times 10^{-6}$ when the number of training iterations is $3/5$  and $4/5$  of the total number of iterations, respectively. The image size of the input model is set to 231×231. batch size is 5.
    
    We propose a completely new dataset for training the network and ablation experiments. The dataset is a random selection of 100,000 images from ImageNet, which ensures the diversity of images. One to four rain patterns are randomly selected from the Efficientderain \cite{ref12} rain pattern dataset and added to the selected images. We finally selected 3000 synthetic images as the training set and 400 as the test set. We named this dataset Rain3000. in addition, we also used the publicly available datasets Rain100L and Rain100H \cite{ref44} to validate the SDnet model. Both publicly available datasets contain 1800 training images and 200 test images.
    
    We use SSIM and PSNR, which have been widely used to evaluate the quality of predicted images, as evaluation metrics. PSNR is calculated based on the pixel error between the two images, and the smaller the error, the larger the value, the more similar the images are, and the better the rain removal effect. On the contrary, the worse the effect of image deraining.
    
    \subsection{Ablation study}
    Our proposed method consists of several key elements, including the package expansion jump connection, the loss function, and the three-branch module Mswt,. We will perform ablation experiments on these three aspects on the dataset Rain3000 to demonstrate the role played by each constituent element.
    
    Table \ref{table1} shows the effect of hopping connections on the network model, where R1 is no hopping connections, R2 is only smaller hopping connections, and R3 is only larger hopping connections. The smaller jump connections improve the PSNR and SSIM by 0.59 dB and 0.0052, respectively, while the larger jump connections improve them by 1.09 dB and 0.0095. Therefore, the fusion of deep and shallow features is still effective for the Swin-transformer, and the larger jump connections are more effective than the smaller ones. Combined with SDNet network, it can be seen that smaller jump connections are less effective based on larger jump connections.
    
        \begin{table}
        \caption{Ablation experiments of jump connections.}
        \label{table1}
        \centering
        \begin{tabular}{@{}ccc@{}}
        \toprule
        Variant & SSIM   & PSNR/dB \\ \midrule
        R1      & 0.9521 & 32.97  \\
        R2      & 0.9573 & 33.56  \\
        R3      & 0.9616 & 34.06  \\
        SDNet   & 0.9621 & 34.09  \\ \bottomrule
        \end{tabular}
        \end{table}
        
        \begin{table}
        \caption{Loss function ablation experiment.}
        \label{table2}
        \centering
        \begin{tabular}{@{}cccccc@{}}
        \toprule
        Loss-Function & SSIM-Loss & L1-Loss & Identity-Loss & SSIM  & PSNR/dB         \\ \midrule
        Loss-1        & $\sqrt{ }$        &         &               & 0.9620 & 34.08          \\
        Loss-2        &           & $\sqrt{ }$      &               & 0.9610 & 34.06          \\
        Loss-3        & $\sqrt{ }$         & $\sqrt{ }$      &               & 0.9620 & 34.08          \\
        SDNet         & $\sqrt{ }$        & $\sqrt{ }$   & $\sqrt{ }$          & 0.9620 & 34.09 \\ \bottomrule
        \end{tabular}
        \end{table}

    Table \ref{table2} shows the performance obtained by training the network using different loss functions. Using (1-SSIM) as the loss function, the network has been able to achieve more satisfactory results, and the identity loss (Identity-Loss) and the absolute deviation sum (L1-Loss) can further improve the performance of the network. This illustrates that the identity loss used for image style conversion is also applicable to the image de-rain task. Table \ref{table3} shows the multi-branch ablation experiments. Multi-branching is inspired by the multi-headed attention mechanism in self-attention. In the table, Mswt-2 is two-branched, Mswt-3 is three-branched, and Mswt-4 is four-branched. Analyzing the results in the table, we can conclude that the more branches there are, the more performance improvement there is and the more parameters there are. To balance the number of parameters and performance, we choose three-branch Mswt-3 as the feature fusion module for SDNet.
        \begin{table}[]
        \caption{Multi-branch ablation experiment.}
        \label{table3}
        \centering
        \begin{tabular}{@{}llll@{}}
        \toprule
        Variant & Parameter & SSIM   & PSNR/dB \\ \midrule
        Mswt-2  & 1.32M     & 0.9603 & 33.86  \\
        Mswt-3  & 1.73M     & 0.9621 & 34.09  \\
        Mswt-4  & 2.14M     & 0.9633 & 34.28  \\ \bottomrule
        \end{tabular}
        \end{table}
    
    \subsection{Comparison experiments}
    
        \begin{table}[]
        \caption{The performance on the Rain3000, Rain100L and Rain100H datasets is indicated by bolded for the best and underlined for the second best.}\label{talbe4}
        \centering
        \begin{tabular}{@{}lllllllll@{}}
        \toprule
        Data Set                  & Evaluation index & \begin{tabular}[c]{@{}l@{}}RESCAN\\    \\ 2018\\    \\ ECCV\end{tabular} & \begin{tabular}[c]{@{}l@{}}NLEDN\\    \\ 2018\\    \\ ACMMM\end{tabular} & \begin{tabular}[c]{@{}l@{}}GCANet\\    \\ 2019\\    \\ WACV\end{tabular} & \begin{tabular}[c]{@{}l@{}}PREnet\\    \\ 2019\\    \\ CVPR\end{tabular} & \begin{tabular}[c]{@{}l@{}}DCSFN\\    \\ 2020\\    \\ ACMMM\end{tabular} & \begin{tabular}[c]{@{}l@{}}MPRnet\\    \\ 2021\\    \\ CVPR\end{tabular} & SDNet            \\ \midrule
        \multirow{2}{*}{Rain3000} & SSIM             & 0.9248                                                                   & 0.9554                                                                   & 0.9350                                                                    & 0.9547                                                                   & {\ul 0.9540}                                                              & 0.9531                                                                   & \textbf{0.9621}  \\
          & PSNR             & 30.76                                                                   & 33.45                                                                   & 31.86                                                                    & 32.82                                                                   & 33.33                                                                   & {\ul 33.83}                                                             & \textbf{34.09}  \\ \midrule
        \multirow{2}{*}{Rain100L} & SSIM             & 0.9629                                                                   & 0.9820                                                                    & 0.9758                                                                   & 0.978                                                                    & {\ul 0.9821}                                                             & 0.9790                                                                    & \textbf{0.9843}  \\
          & PSNR             & 33.99                                                                    & 37.12                                                                   & 36.18                                                                    & 35.94                                                                    & 37.60                                                                   & {\ul 37.81}                                                              & \textbf{37.92} \\ \midrule
        \multirow{2}{*}{Rain100H} & SSIM             & 0.7612                                                                   & 0.8650                                                                    & 0.8200                                                                     & 0.8661                                                                   & {\ul 0.8860}                                                              & 0.8460                                                                    & \textbf{0.8957}  \\
          & PSNR             & 23.98                                                                    & 27.21                                                                    & 25.99                                                                    & 26.10                                                                   & {\ul 27.76}                                                              & 26.31                                                                    & \textbf{28.26}  \\ \bottomrule
        \end{tabular}
        \end{table}
        
        \begin{figure}
        \caption{Algorithm comparison chart}
        \centering
        \includegraphics[scale=0.5]{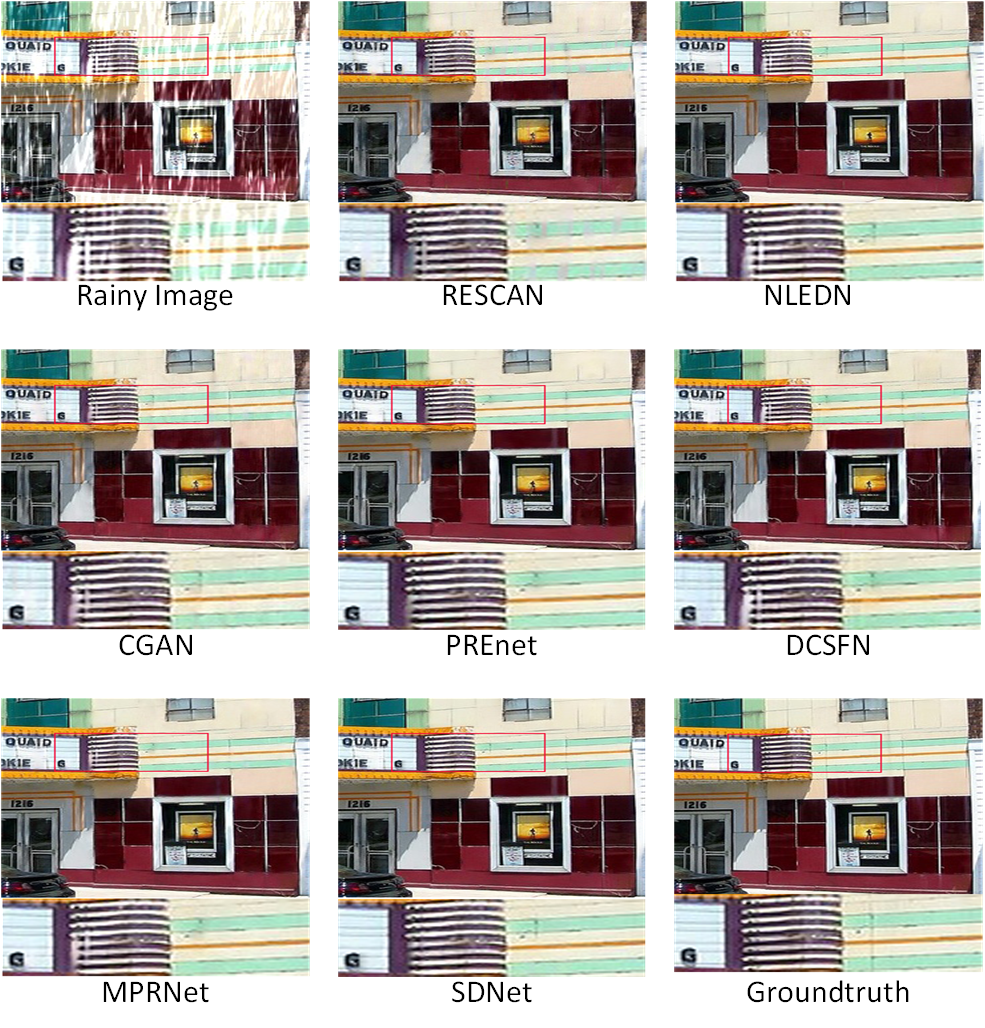}
        \label{fig6}
        \end{figure}
    Table \ref{talbe4} shows our approach compared with six state-of-the-art methods, including the contextual aggregation network with SE (Squeeze-and-Excitation) module (RESCAN) \cite{ref45}, the fused high and low level features defogging and rain removal network (GCANet) \cite{ref46}, the non-locally enhanced self-coding network (NLEDN) \cite{ref47}, the multi-stage progressive in rain network (PREnet) \cite{ref41}, multi-scale fusion network (DCSFN) \cite{ref48} and the latest pair of phase progressive restoration network (MPRnet) \cite{ref49}. The experimental results show that our proposed SDNet network has the most significant advantage over the 2018 RESCAN network in terms of evaluation metrics PSNR and SSIM on the dataset Rain100H, with an improvement of 4.28 dB and 0.1345, respectively. compared to the latest MPRNet model, PSNR and SSIM improve by 1.95 dB and 0.0497, respectively. Compared with the next best DCSFN, the PSNR and SSIM of our proposed model improve 0.5 dB and 0.0097 on Rain100H, respectively. therefore, our method outperforms the above six models.
    
    Figure \ref{fig6} shows the visual effect of rain removal by each algorithm. It is clear from the results that RESCAN has artifacts, NLEDN, CGAN, DCSFN, and MPRNet have good rain removal, but some rain patterns are not removed, and PREnet can remove rain patterns, but some texture details in the background are also removed. These six models have some shortcomings in the recovery effect, while SDNet can both remove the rain pattern well and recover the texture details satisfactorily.

        \begin{figure}[htbp]
        \begin{minipage}[t]{0.5\linewidth}
        \centering
        \includegraphics[height=5.5cm,width=7cm]{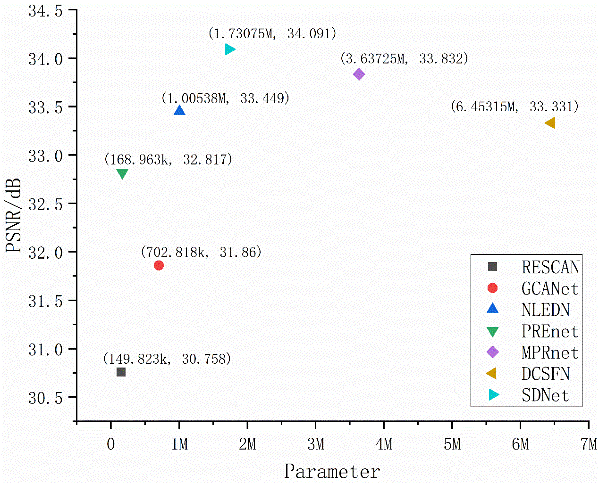}
        \caption{Parameter comparison graph}
        \label{fig7}
        \end{minipage}%
        \hfill
        \begin{minipage}[t]{0.5\linewidth}
        \centering
        \includegraphics[height=5.5cm,width=7cm]{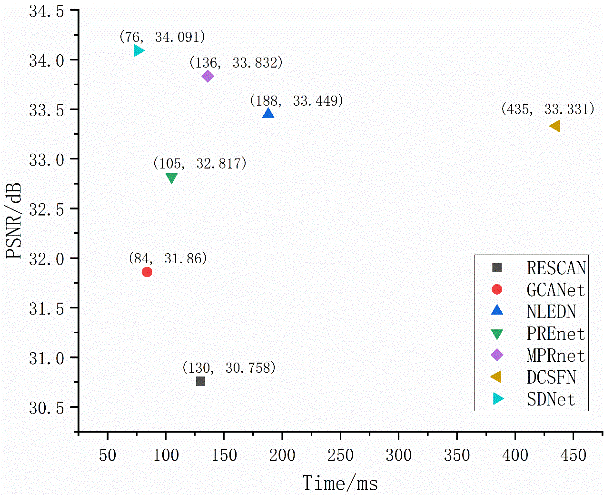}
        \caption{Time consumption graph}
        \label{fig8}
        \end{minipage}
        \end{figure}
    
    Figure \ref{fig7} shows the comparison of the number of model parameters and performance. It can be seen from the figure that the number of parameters of SDPNet is much smaller than that of DCSFN and MPRNet, only 26.82$\%$ of DCSFN and 47.58$\%$ of MPRNet. Figure \ref{fig8} shows the comparison of model inference time and model performance. From the figure, it can be seen that the inference speed of SDNet is the fastest, 5.7 times faster than DCSFN and 1.79 times faster than the latest model MPRNet. Combining the two comparison plots, SDNet can achieve the fastest inference and optimal deraining performance without a significant increase in the number of parameters, which further demonstrates the practical value of our proposed model.

\section{Conclusion and Prospect}
    First, we propose a three-branch end-to-end rain removal network based on Swin-transformer, which makes full use of the powerful learning ability of the transformer by replacing the convolution operation with a modified Swin-transformer, and in addition we design a multi-branch module to fuse information from different spatial domains, using jump connections to fuse deep features and shallow features. In addition, we propose a new dataset which consists of 3000 training pairs and 400 test pairs. The dataset is produced based on ImageNet with a rich combination of background and rain patterns, which facilitates the generalization of the model. Our proposed model achieves optimal performance on both the dataset Rain3000 and the public datasets Rain100L, Rain100H.
    
    Our work still has some shortcomings. For example, it is not explored in detail which approach is more suitable for the image deraining task, parallel or serial, with the same number of parameters. And whether it is possible to use multiple sliding windows of different sizes to achieve further information exchange between windows to improve the performance of network deraining. In addition, we are using simpler feedforward networks,
    
    and more complex networks are still worth investigating.

\bibliographystyle{unsrt}
\bibliography{references}

\begin{thebibliography}{10}

\bibitem{ref1}
Liang-Jian Deng, Ting-Zhu Huang, Xi-Le Zhao, and Tai-Xiang Jiang.
\newblock A directional global sparse model for single image rain removal.
\newblock {\em Applied Mathematical Modelling}, 59:662--679, 2018.

\bibitem{ref2}
Li-Wei Kang, Chia-Wen Lin, and Yu-Hsiang Fu.
\newblock Automatic single-image-based rain streaks removal via image
  decomposition.
\newblock {\em IEEE transactions on image processing}, 21(4):1742--1755, 2011.

\bibitem{ref3}
Yinglong Wang, Shuaicheng Liu, Chen Chen, and Bing Zeng.
\newblock A hierarchical approach for rain or snow removing in a single color
  image.
\newblock {\em IEEE Transactions on Image Processing}, 26(8):3936--3950, 2017.

\bibitem{ref4}
Yu~Li, Robby~T Tan, Xiaojie Guo, Jiangbo Lu, and Michael~S Brown.
\newblock Rain streak removal using layer priors.
\newblock In {\em Proceedings of the IEEE conference on computer vision and
  pattern recognition}, pages 2736--2744, 2016.

\bibitem{ref5}
Lei Zhu, Chi-Wing Fu, Dani Lischinski, and Pheng-Ann Heng.
\newblock Joint bi-layer optimization for single-image rain streak removal.
\newblock In {\em Proceedings of the IEEE international conference on computer
  vision}, pages 2526--2534, 2017.

\bibitem{ref6}
Yu~Luo, Yong Xu, and Hui Ji.
\newblock Removing rain from a single image via discriminative sparse coding.
\newblock In {\em Proceedings of the IEEE International Conference on Computer
  Vision}, pages 3397--3405, 2015.

\bibitem{ref7}
Hong Wang, Qi~Xie, Qian Zhao, and Deyu Meng.
\newblock A model-driven deep neural network for single image rain removal.
\newblock In {\em Proceedings of the IEEE/CVF Conference on Computer Vision and
  Pattern Recognition}, pages 3103--3112, 2020.

\bibitem{ref8}
Cong Wang, Yutong Wu, Zhixun Su, and Junyang Chen.
\newblock Joint self-attention and scale-aggregation for self-calibrated
  deraining network.
\newblock In {\em Proceedings of the 28th ACM International Conference on
  Multimedia}, pages 2517--2525, 2020.

\bibitem{ref9}
Xueyang Fu, Qi~Qi, Zheng-Jun Zha, Yurui Zhu, and Xinghao Ding.
\newblock Rain streak removal via dual graph convolutional network.
\newblock In {\em Proc. AAAI Conf. Artif. Intell}, pages 1--9, 2021.

\bibitem{ref10}
Nanfeng Jiang, Weiling Chen, Liqun Lin, and Tiesong Zhao.
\newblock Single image rain removal via multi-module deep grid network.
\newblock {\em Computer Vision and Image Understanding}, 202:103106, 2021.

\bibitem{ref11}
Yiyang Shen, Yidan Feng, Sen Deng, Dong Liang, Jing Qin, Haoran Xie, and
  Mingqiang Wei.
\newblock Mba-raingan: Multi-branch attention generative adversarial network
  for mixture of rain removal from single images.
\newblock {\em arXiv preprint arXiv:2005.10582}, 2020.

\bibitem{ref12}
Qing Guo, Jingyang Sun, Felix Juefei-Xu, Lei Ma, Xiaofei Xie, Wei Feng, and
  Yang Liu.
\newblock Efficientderain: Learning pixel-wise dilation filtering for
  high-efficiency single-image deraining.
\newblock {\em arXiv preprint arXiv:2009.09238}, 2020.

\bibitem{ref13}
Yeachan Park, Myeongho Jeon, Junho Lee, and Myungjoo Kan.
\newblock Mara-net: Single image deraining network with multi-level connection
  and adaptive regional attention.
\newblock {\em arXiv preprint arXiv:2009.13990}, 2020.

\bibitem{ref14}
Cong Wang, Wanshu Fan, Honghe Zhu, and Zhixun Su.
\newblock Single image deraining via nonlocal squeeze-and-excitation enhancing
  network.
\newblock {\em Applied Intelligence}, 50(9):2932--2944, 2020.

\bibitem{ref15}
Cong Wang, Yutong Wu, Yu~Cai, Guangle Yao, Zhixun Su, and Hongyan Wang.
\newblock Single image deraining via deep pyramid network with spatial
  contextual information aggregation.
\newblock {\em Applied Intelligence}, pages 1--11, 2020.

\bibitem{ref16}
Wenhan Yang, Robby~T Tan, Shiqi Wang, Yuming Fang, and Jiaying Liu.
\newblock Single image deraining: From model-based to data-driven and beyond.
\newblock {\em IEEE Transactions on pattern analysis and machine intelligence},
  2020.

\bibitem{ref17}
Siyuan Li, Wenqi Ren, Jiawan Zhang, Jinke Yu, and Xiaojie Guo.
\newblock Single image rain removal via a deep decomposition--composition
  network.
\newblock {\em Computer Vision and Image Understanding}, 186:48--57, 2019.

\bibitem{ref27}
Yi-Lei Chen and Chiou-Ting Hsu.
\newblock A generalized low-rank appearance model for spatio-temporally
  correlated rain streaks.
\newblock In {\em Proceedings of the IEEE International Conference on Computer
  Vision}, pages 1968--1975, 2013.

\bibitem{ref18}
Cong Wang, Man Zhang, Zhixun Su, Yutong Wu, Guangle Yao, and Hongyan Wang.
\newblock Learning a multi-level guided residual network for single image
  deraining.
\newblock {\em Signal Processing: Image Communication}, 78:206--215, 2019.

\bibitem{ref19}
Sen Deng, Mingqiang Wei, Jun Wang, Yidan Feng, Luming Liang, Haoran Xie, Fu~Lee
  Wang, and Meng Wang.
\newblock Detail-recovery image deraining via context aggregation networks.
\newblock In {\em Proceedings of the IEEE/CVF Conference on Computer Vision and
  Pattern Recognition}, pages 14560--14569, 2020.

\bibitem{ref20}
Yanyan Wei, Zhao Zhang, Haijun Zhang, Jie Qin, and Mingbo Zhao.
\newblock Semi-deraingan: A new semi-supervised single image deraining network.
\newblock {\em arXiv preprint arXiv:2001.08388}, 2020.

\bibitem{ref21}
Rajeev Yasarla, VA~Sindagi, and VM~Patel.
\newblock Semi-supervised image deraining using gaussian processes.
\newblock {\em arXiv preprint arXiv:2009.13075}, 2020.

\bibitem{ref22}
Wei Wei, Deyu Meng, Qian Zhao, Zongben Xu, and Ying Wu.
\newblock Semi-supervised transfer learning for image rain removal.
\newblock In {\em Proceedings of the IEEE/CVF Conference on Computer Vision and
  Pattern Recognition}, pages 3877--3886, 2019.

\bibitem{ref34}
Rajeev Yasarla, Vishwanath~A Sindagi, and Vishal~M Patel.
\newblock Syn2real transfer learning for image deraining using gaussian
  processes.
\newblock In {\em Proceedings of the IEEE/CVF Conference on Computer Vision and
  Pattern Recognition}, pages 2726--2736, 2020.

\bibitem{ref35}
Kewen Han and Xinguang Xiang.
\newblock Decomposed cyclegan for single image deraining with unpaired data.
\newblock In {\em ICASSP 2020-2020 IEEE International Conference on Acoustics,
  Speech and Signal Processing (ICASSP)}, pages 1828--1832. IEEE, 2020.

\bibitem{ref36}
Yanyan Wei, Zhao Zhang, Jicong Fan, Yang Wang, Shuicheng Yan, and Meng Wang.
\newblock Derain-cyclegan: An attention-guided unsupervised benchmark for
  single image deraining and rainmaking.
\newblock {\em arXiv preprint arXiv:1912.07015}, 2019.

\bibitem{ref37}
Hongyuan Zhu, Xi~Peng, Joey~Tianyi Zhou, Songfan Yang, Vijay Chanderasekh,
  Liyuan Li, and Joo-Hwee Lim.
\newblock Singe image rain removal with unpaired information: A differentiable
  programming perspective.
\newblock In {\em Proceedings of the AAAI Conference on Artificial
  Intelligence}, volume~33, pages 9332--9339, 2019.

\bibitem{ref23}
Ashish Vaswani, Noam Shazeer, Niki Parmar, Jakob Uszkoreit, Llion Jones,
  Aidan~N Gomez, Lukasz Kaiser, and Illia Polosukhin.
\newblock Attention is all you need.
\newblock {\em arXiv preprint arXiv:1706.03762}, 2017.

\bibitem{ref24}
Alexey Dosovitskiy, Lucas Beyer, Alexander Kolesnikov, Dirk Weissenborn,
  Xiaohua Zhai, Thomas Unterthiner, Mostafa Dehghani, Matthias Minderer, Georg
  Heigold, Sylvain Gelly, et~al.
\newblock An image is worth 16x16 words: Transformers for image recognition at
  scale.
\newblock {\em arXiv preprint arXiv:2010.11929}, 2020.

\bibitem{ref25}
Daquan Zhou, Bingyi Kang, Xiaojie Jin, Linjie Yang, Xiaochen Lian, Zihang
  Jiang, Qibin Hou, and Jiashi Feng.
\newblock Deepvit: Towards deeper vision transformer.
\newblock {\em arXiv preprint arXiv:2103.11886}, 2021.

\bibitem{ref26}
Chun-Fu Chen, Quanfu Fan, and Rameswar Panda.
\newblock Crossvit: Cross-attention multi-scale vision transformer for image
  classification.
\newblock {\em arXiv preprint arXiv:2103.14899}, 2021.

\bibitem{ref29}
Ze~Liu, Yutong Lin, Yue Cao, Han Hu, Yixuan Wei, Zheng Zhang, Stephen Lin, and
  Baining Guo.
\newblock Swin transformer: Hierarchical vision transformer using shifted
  windows.
\newblock {\em arXiv preprint arXiv:2103.14030}, 2021.

\bibitem{ref28}
Xueyang Fu, Jiabin Huang, Xinghao Ding, Yinghao Liao, and John Paisley.
\newblock Clearing the skies: A deep network architecture for single-image rain
  removal.
\newblock {\em IEEE Transactions on Image Processing}, 26(6):2944--2956, 2017.

\bibitem{ref30}
Yingjun Du, Jun Xu, Xiantong Zhen, Ming-Ming Cheng, and Ling Shao.
\newblock Conditional variational image deraining.
\newblock {\em IEEE Transactions on Image Processing}, 29:6288--6301, 2020.

\bibitem{ref31}
He~Zhang and Vishal~M Patel.
\newblock Density-aware single image de-raining using a multi-stream dense
  network.
\newblock In {\em Proceedings of the IEEE conference on computer vision and
  pattern recognition}, pages 695--704, 2018.

\bibitem{ref32}
Jingwei He, Lei Yu, Gui-Song Xia, and Wen Yang.
\newblock Single image deraining with continuous rain density estimation.
\newblock {\em arXiv preprint arXiv:2006.03190}, 2020.

\bibitem{ref33}
Kui Jiang, Zhongyuan Wang, Peng Yi, Chen Chen, Baojin Huang, Yimin Luo, Jiayi
  Ma, and Junjun Jiang.
\newblock Multi-scale progressive fusion network for single image deraining.
\newblock In {\em Proceedings of the IEEE/CVF Conference on Computer Vision and
  Pattern Recognition}, pages 8346--8355, 2020.

\bibitem{ref38}
Jieneng Chen, Yongyi Lu, Qihang Yu, Xiangde Luo, Ehsan Adeli, Yan Wang, Le~Lu,
  Alan~L Yuille, and Yuyin Zhou.
\newblock Transunet: Transformers make strong encoders for medical image
  segmentation.
\newblock {\em arXiv preprint arXiv:2102.04306}, 2021.

\bibitem{ref39}
Yifan Jiang, Shiyu Chang, and Zhangyang Wang.
\newblock Transgan: Two transformers can make one strong gan.
\newblock {\em arXiv preprint arXiv:2102.07074}, 2021.

\bibitem{ref40}
Ren{\'e} Ranftl, Alexey Bochkovskiy, and Vladlen Koltun.
\newblock Vision transformers for dense prediction.
\newblock {\em arXiv preprint arXiv:2103.13413}, 2021.

\bibitem{ref41}
Dongwei Ren, Wangmeng Zuo, Qinghua Hu, Pengfei Zhu, and Deyu Meng.
\newblock Progressive image deraining networks: A better and simpler baseline.
\newblock In {\em Proceedings of the IEEE/CVF Conference on Computer Vision and
  Pattern Recognition}, pages 3937--3946, 2019.

\bibitem{ref42}
Jun-Yan Zhu, Taesung Park, Phillip Isola, and Alexei~A Efros.
\newblock Unpaired image-to-image translation using cycle-consistent
  adversarial networks.
\newblock In {\em Proceedings of the IEEE international conference on computer
  vision}, pages 2223--2232, 2017.

\bibitem{ref43}
Diederik~P Kingma and Jimmy Ba.
\newblock Adam: A method for stochastic optimization.
\newblock {\em arXiv preprint arXiv:1412.6980}, 2014.

\bibitem{ref44}
Wenhan Yang, Robby~T Tan, Jiashi Feng, Jiaying Liu, Zongming Guo, and Shuicheng
  Yan.
\newblock Deep joint rain detection and removal from a single image.
\newblock In {\em Proceedings of the IEEE Conference on Computer Vision and
  Pattern Recognition}, pages 1357--1366, 2017.

\bibitem{ref45}
Xia Li, Jianlong Wu, Zhouchen Lin, Hong Liu, and Hongbin Zha.
\newblock Recurrent squeeze-and-excitation context aggregation net for single
  image deraining.
\newblock In {\em Proceedings of the European Conference on Computer Vision
  (ECCV)}, pages 254--269, 2018.

\bibitem{ref46}
Dongdong Chen, Mingming He, Qingnan Fan, Jing Liao, Liheng Zhang, Dongdong Hou,
  Lu~Yuan, and Gang Hua.
\newblock Gated context aggregation network for image dehazing and deraining.
\newblock In {\em 2019 IEEE winter conference on applications of computer
  vision (WACV)}, pages 1375--1383. IEEE, 2019.

\bibitem{ref47}
Guanbin Li, Xiang He, Wei Zhang, Huiyou Chang, Le~Dong, and Liang Lin.
\newblock Non-locally enhanced encoder-decoder network for single image
  de-raining.
\newblock In {\em Proceedings of the 26th ACM international conference on
  Multimedia}, pages 1056--1064, 2018.

\bibitem{ref48}
Cong Wang, Xiaoying Xing, Yutong Wu, Zhixun Su, and Junyang Chen.
\newblock Dcsfn: Deep cross-scale fusion network for single image rain removal.
\newblock In {\em Proceedings of the 28th ACM International Conference on
  Multimedia}, pages 1643--1651, 2020.

\bibitem{ref49}
Syed~Waqas Zamir, Aditya Arora, Salman Khan, Munawar Hayat, Fahad~Shahbaz Khan,
  Ming-Hsuan Yang, and Ling Shao.
\newblock Multi-stage progressive image restoration.
\newblock {\em arXiv preprint arXiv:2102.02808}, 2021.

\end{thebibliography}

\end{document}